\documentclass[letterpaper, 10 pt, conference]{ieeeconf}  

\IEEEoverridecommandlockouts                              

\usepackage{times} 
\usepackage{amsmath} 
\usepackage{amssymb}  
\usepackage{algpseudocode,algorithm,algorithmicx}
\usepackage[keeplastbox]{flushend} 
\usepackage{float}
\usepackage{subfig}
\usepackage{graphicx}
\usepackage{multirow}
\usepackage{bm}
\usepackage{booktabs}
\usepackage{array}
\usepackage{gensymb} 
\usepackage{threeparttable}
\usepackage{algpseudocode,algorithm,algorithmicx}
\usepackage[dvipsnames]{xcolor}
\usepackage{cite}
\definecolor{red}{rgb}{1, 0, 0}
\definecolor{green}{rgb}{0, 1, 0}
\definecolor{grey}{rgb}{0.8,0.8,0.8}
\definecolor{yellow}{rgb}{1,1,0}
\newcolumntype{C}[1]{>{\centering}m{#1}}
\algnewcommand\True{\textbf{true}\space}
\newcommand{\Map}[1]{\mathbb{M}_{\text{#1}}}
\newcommand{\norm}[1]{\left\lVert{#1}\right\rVert}

\title{\LARGE \bf
Robotic Exploration of Unknown 2D Environment Using a Frontier-based Automatic-Differentiable Information Gain Measure
}

\author{Di Deng, Runlin Duan, Jiahong Liu, Kuangjie Sheng, and Kenji Shimada
\thanks{Department of Mechanical Engineering, Carnegie Mellon University, 5000 Forbes Ave, Pittsburgh, PA 15213, USA.,
		{\tt\small dengd@andrew.cmu.edu}}
}

\begin{document}

\maketitle
\thispagestyle{empty}
\pagestyle{empty}

\noindent \begin{abstract} 
At the heart of path-planning methods for autonomous robotic exploration is a heuristic which encourages exploring unknown regions of the environment. Such heuristics are typically computed using frontier-based or information-theoretic methods. Frontier-based methods define the information gain of an exploration path as the number of boundary cells, or frontiers, which are visible from the path. However, the discrete and non-differentiable nature of this measure of information gain makes it difficult to optimize using gradient-based methods. In contrast, information-theoretic methods define information gain as the mutual information between the sensor's measurements and the explored map. However, computation of the gradient of mutual information involves finite differencing and is thus computationally expensive. In this work, we propose an exploration planning framework which combines sampling-based path planning and gradient-based path optimization. The main contribution of this framework is a novel reformulation of information gain as a differentiable function. This allows us to simultaneously optimize information gain with other differentiable quality measures, such as smoothness. The effectiveness of the proposed planning framework is verified both in simulation and in hardware experiments using a Turtlebot3 Burger robot.

\end{abstract}

\section{Introduction}
\begin{figure}[t]
    \centering
    \includegraphics[width = 0.99\linewidth]{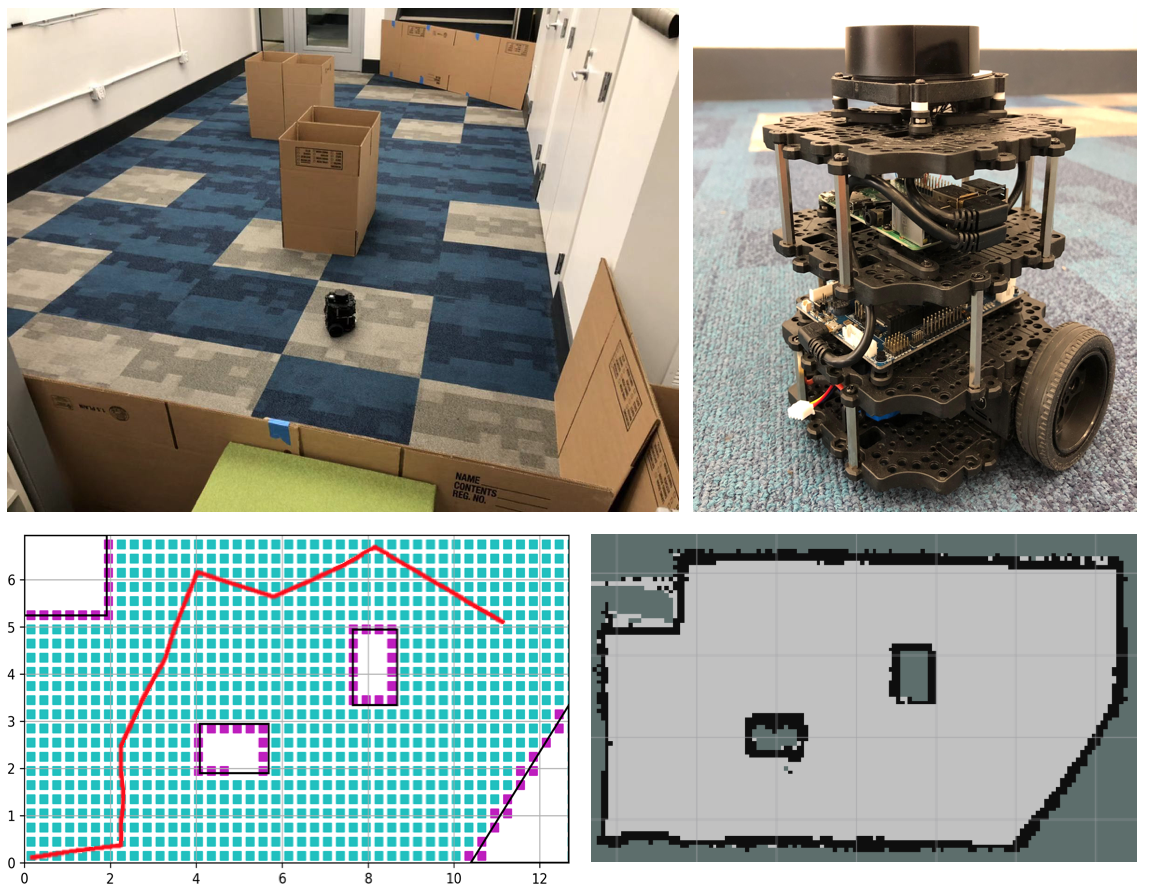}
    \caption{Path planning for area coverage with a LIDAR-equiped robot.}
    \label{fig.intro}
    \vspace{-0.3cm}
\end{figure}

\noindent There has been an increasing demand for exploring both indoor and outdoor areas using autonomous vehicles equipped with range sensors such as LIDARs and depth cameras. Typical applications of area exploration include search and surveillance, mapping, lawn mowing, floor cleaning, etc. These applications require robots that generate paths automatically covering bounded regions based on onboard sensor readings.  Fig. \ref{fig.intro} shows an area exploration task for exploring an indoor environment with a Turtlebot3 Burger robot.

{}

To explore an unknown region, the region is typically represented as a probabilistic occupancy map containing cells. The probability of each cell represents whether the cell is unknown, free, or occupied. As the robot explores the region, the occupancy probability is updated with the robot's sensor measurements.

Guiding the robot to exhaustively and efficiently cover the entire unknown space is most commonly carried out in a receding-horizon fashion: the robot plans a path, follows the path, updates the map with the sensor data collected along the path, and repeats this process until the region is fully explored \cite{bircher2016receding}. Existing approaches typically search for exploration paths using a collision-free path planner \cite{lavalle2006planning}, and then try to optimize the information gain of an exploration path by finding the view-point that is most likely to maximize exploring unknown regions \cite{song2017online, song2018surface, blaer2007data}. Two main approaches have been proposed to quantify the information gain of a path: (\textbf{i}) frontier-based \cite{yamauchi1998frontier, holz2010evaluating} and (\textbf{ii}) information-theoretic methods \cite{julian2014mutual, bai2016information, amigoni2010information}. 

Frontier-based methods measure information gain by counting the number of boundary cells observable from the path. As it is a non-differentiable function of the path, the optimization of information gain is typically achieved by computing the information gain for a finite number of samples and choosing one with the largest information gain. 

On the other hand, information-theoretic methods are designed to optimize an objective function such as mutual information or Shannon's entropy \cite{ bourgault2002information, stachniss2005information}. However, information-theoretic methods such as \cite{charrow2015information} denote information gain as a function of a discrete set of robot positions and compute gradients using finite differences. Both metrics are computationally expensive, limiting the replanning frequency of a robot during exploration.

\textbf{Contribution.}  We introduce a novel algorithm for the exploration problem with a robot for area exploration tasks. The proposed approach reformulates a frontier-based information gain measure as an automatic-differentiable function with respect to the robot's path, allowing the path to be optimized with gradient-based optimization. In contrast, traditional frontier-based methods typically optimize information gain by sampling a large number of paths. In addition, compared with information-theoretic methods, gradient computed with the proposed automatic differentiation approach has higher accuracy and less computational complexity.  

\section{Related Work \label{sec.related}}
\noindent \textbf{Frontier-based methods.} Connolly defines frontiers as the region which separates the free space and unexplored regions of a grid map and evaluates the information gain of a view-point as the number of frontier cells visible from a view-point \cite{connolly1985determination}. The straightforward use of frontiers to encourage exploration is to navigate a robot directly to the frontier cells  \cite{yamauchi1997frontier}. Freda \textit{et al.} \cite{freda2005frontier} use a frontier-based modification of the Sensor-based Random Tree (SRT) method that only samples configurations within a maximum collision-free circular or star-shaped local safe region around a robot. By deploying a probabilistic strategy, the randomly generated configurations are biased towards frontier regions. Dornhege \textit{et al.} \cite{dornhege2013frontier} proposed a more sophisticated strategy for information gain measurement by combining frontiers with void cells, which are defined as unknown cells within the convex hull of occupied cells.

Finding the view-point of the maximum information gain is known as the Next Best View (NBV) problem. This view-point is determined by sampling a significant number of candidate view-points, counting the frontier observed by each view-point, and selecting the one with the highest information gain \cite{bircher2016three}. Instead of searching a single view-point, Bircher \textit{et al.} propose a receding-horizon approach that finds paths as the best branch in a rapidly exploring random tree (RRT)\cite{lavalle2006planning}. To extend from evaluating the vertices of the path, Song and Jo maximize information gain along the entire inspection path by iteratively reducing sampling range and employing the streaming set cover algorithm \cite{song2017online}.

\textbf{Information-theoretic methods} maximize the mutual information between the discrete occupancy map and the future sensor's measurement. Julian \textit{et al.} prove that by maximizing Shannon's mutual information \cite{verdu1998fifty}, a robot is guaranteed to explore unknown space. However, Nelson and Michael \cite{nelson2015information} point out this approach limits the control frequency. To be computationally more efficient, Charrow \textit{et al.} \cite{charrow2015informationmapping} utilize the Cauchy-Schwarz quadratic mutual information (CSQMI) \cite{charrow2015information} based on Renyi's quadratic entropy (RQE) \cite{renyi1976some} for selecting paths rather than destinations. CSQMI is substantially faster, but it is still computationally demanding due to the finite differences involved in computing the gradient of the mutual information. 

\textbf{Planning a path to the goal.} After the aforementioned algorithms find a goal, a path from the robot's current configuration to the goal can be found by either a sampling-based method such as RRT \cite{lavalle2006planning}, or a trajectory-optimization method \cite{schulman2013finding, ratliff2009chomp} such as CHOMP. Sampling-based methods have the advantage of achieving a feasible solution while avoiding local minima \cite{hollinger2014sampling}; optimization-based approaches perform better at locally optimizing the information gain and path length.

\begin{figure*}[t]
\vspace{2mm}
    \centering
    \includegraphics[width=1.0\linewidth]{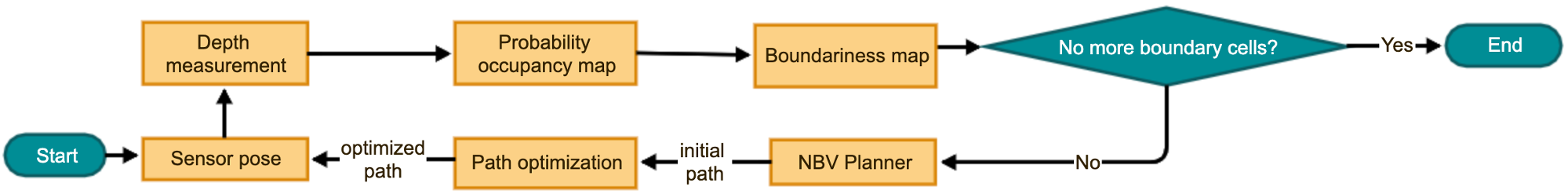}
    \caption{Overview of the proposed optimization-based algorithm for coverage planning}
    \label{fig:overview}
\end{figure*}
  
\vspace{-0.1cm}
\section{Overview \label{sec.overview}}
\noindent The goal of the proposed planning framework is to explore an unknown bounded 2D region, $\textbf{V} \subset \mathbb{R}^2$, with a ground robot to determine the subsets of $\textbf{V}$, which are free ($\textbf{V}_{\text{free}} \subseteq \textbf{V}$) and occupied ($\textbf{V}_{\text{occ}} \subseteq \textbf{V}$). Region $\textbf{V}$ is represented by a probabilistic occupancy grid map, $\Map{prob} \in \mathbb{R}^{m \times n}$, with cell resolution $\varrho$. The $(i, j)$ entry of $\Map{prob}$, denoted by $\Map{prob}(i,j) \in [0, 1]$,  is the probability of the cell at $(i,j)$ being occupied, free, or unknown:
\begin{equation}
    \Map{prob}(i,j) = \begin{cases}
  \text{free} &  < 0.5 \\    
  \text{unknown} &  = 0.5 \\
  \text{occupied}  &  > 0.5
\end{cases}.
\end{equation}

The robot starts at an initial configuration, or view-point, denoted by $\xi_0 = [x, y, \theta]$. The planning framework is supposed to to plan collision-free paths iteratively, $\Xi = \{\xi_0, \dots, \xi_n\}$, and use the sensor measurements along the paths to update $\Map{prob}$ until the number of unknown cells becomes smaller than a specified threshold.

An overview of our optimization-based coverage planning framework is given in Fig. \ref{fig:overview}. The process starts from a collision-free initial position, perceives the environment using a 2D range sensor, and updates $\Map{prob}$ accordingly. Then, a sampling-based NBV planner is adopted to generate a piecewise linear path within the explored free space, $\textbf{V}_{free}$. The smoothness and information gain of the path generated by the sampling-based planner is then optimized using a differentiable objective function. These steps are executed repeatedly until a termination criterion is satisfied.

\section{Probabilistic Occupancy Map}
\noindent This section reviews the range sensor model and the classical probabilistic sensor fusion algorithm used in this paper.

\subsection{2D Range-Sensor Model\label{subsec.camera_model}}
\begin{figure}[htp!]
    \centering
    \includegraphics[width = 0.45\linewidth]{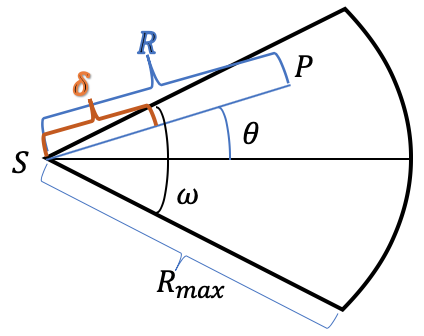}
    \caption{Sensor model; $R_{\text{max}}$ is the maximum range measurement, $\omega$ is the beam aperture, $\textbf{S}$ and $\textbf{P}$ are the positions of sensor and obstacle, R is depth measurement, $\theta$ is the angle between the sensor's central axis and \textbf{SP}, and $\delta$ is the distance of a cell to sensor along $\textbf{SP}$.}
    \label{fig:sensor_model}
    \vspace{-0.3cm}
\end{figure}

\noindent The range sensor shown in Fig. \ref{fig:sensor_model} has a maximum range $R_{\text{max}}$, beam aperture $\omega$, and mean deviation error $\epsilon$. The probability of whether a cell is occupied or free can be computed from range sensor measurements using the model proposed by Moravec and Elfes \cite{moravec1985high}:
\begin{equation}
\label{eqn.Er}
      O_r = \begin{cases}
  1 - (\frac{\delta}{R-\epsilon})^2 &  \delta \in [0, R - \epsilon]\\    
  (\frac{\delta - R}{\epsilon})^2 - 1 &  \delta \in [R - \epsilon, R + \epsilon]
\end{cases},
\end{equation}

\begin{equation}
\label{eqn.PO}
  O_a = 1 - (\frac{2\theta}{\omega})^2, \rho_O= 
    \frac{1 + O_r(\delta)O_a(\theta)}{2}, \theta \in [-\frac{\omega}{2}, \frac{\omega}{2}],
\end{equation}
where $O_r, O_a$ and $\rho_O$ are the distance, angle, and overall probability that a cell with distance $\delta$ along ray $\textbf{SP}$ is occupied.

\subsection{Binary Bayes Filter (BBF) \label{subsec.binary_bayes_filter}}
\noindent BBF \cite{thrun2005probabilistic} updates its belief of whether each cell is empty, free, or unknown by fusing range sensor measurements, $z_{t}$, over time. Instead of directly dealing with probabilities, BBF uses the \textit{log-odds} of probabilities, which is defined by:
\begin{equation}
    \label{eqn:log_odds}
    \Map{odds} = \log\frac{\Map{prob}}{1 - \Map{prob}}
\end{equation}
such that: 
\begin{equation}
    \mathbb{M}_{\text{odds}}(i,j) = \begin{cases}
    < 0 & \text{free} \\
    \log\frac{0.5}{1-0.5}=0 & \text{unknown} \\
    > 0 & \text{occupied}
    \end{cases}.
\end{equation}

Since there is a one-to-one mapping between probability and log-odds, the two representations of a probabilistic occupancy map are equivalent. The advantage of log-odds is that a Bayesian update can be implemented simply as the addition between the previous belief, $\mathbb{M}_{\text{prob}}(i,j|z_{1:t-1})$, and the update, $\Map{odds}(i,j|z_t)$. To improve numerical stability and adapt to dynamic environments with moving obstacles, upper and lower bounds, $l_{\text{min}}$ and $l_{\text{max}}$, on log-odds are added during the updates \cite{yguel2008update}. The adjusted update rule is given by:
\begin{equation}
    \begin{array}{l}
         \mathbb{M}_{\text{odds}}(i,j|z_{1:t}) =  \\
         \max(\min(\mathbb{M}_{\text{odds}}(i,j|z_{1:t-1}) + \Map{odds}(i,j|z_t), l_{\text{max}}),l_{\text{min}}).
    \end{array}
\end{equation}

\vspace{-0.2cm}
\section{Boundariness Map}
\noindent A \textit{boundary} cell of a probabilistic occupancy map is classically defined as an unknown cell with at least one neighboring free cell. It represents the boundary between explored and unexplored regions and has been used as heuristics to guide exploration. For example, a common way to compute the information gain of a view-point, $\xi$, is to count the number of boundary cells observable by a robot at $\xi$ \cite{song2017online}. Computer vision algorithms, such as edge detection, can also be applied to detect boundaries \cite{yamauchi1997frontier}. 

\begin{figure}[htp!]
    \includegraphics[width = 0.98\linewidth]{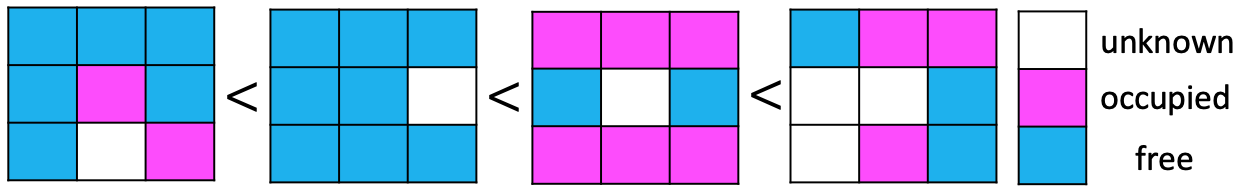}
    \caption{Boundariness of the central cell computed from its neighbors.}
    \label{fig.bd_score}
\end{figure}

In this section, we define a numerical measure for cells called the \textit{boundariness}. It is a number between 0 and 1. The closer a cell's boundariness is to 1, the more likely it is to be a boundary cell. The boundariness of Cell $(i,j)$ is computed from the log-odds of itself and its eight neighboring cells: 
\begin{equation}
\label{eqn.sum_nei}
    p = \sum_{(i, j)\in \textbf{S}}\mathbb{M}_{\text{odds}}(i_0+i,j_0+j),
\end{equation}{}
\begin{equation}
\label{eqn.boundariness}
 \Map{bd}(i_0, j_0) = w\exp{\frac{-\Map{odds}(i_0,j_0)^2}{2\sigma^2}} + (1-w)\exp{\frac{-p^2}{2|\textbf{S}|^2\sigma^2}},
\end{equation}
where $\textbf{S}=\{(\pm1,\pm1),(\pm1, \mp1),(\pm1,0),(0,\pm1)\}\in\mathbb{R}^2$ is the set of relative indices of neighboring cells, and $w$ a constant. The first term in Eqn. (\ref{eqn.boundariness}) rewards the fact that the cell is unknown. The second term is based on the observation that the cell is more likely to be a boundary if its neighboring cells contain a uniform mixture of occupied and free cells, e.g., four occupied and four free. Some examples of this heuristic are shown in Fig. \ref{fig.bd_score}.

\begin{figure}[htp!]
    \centering
    \includegraphics[width = 0.95\linewidth]{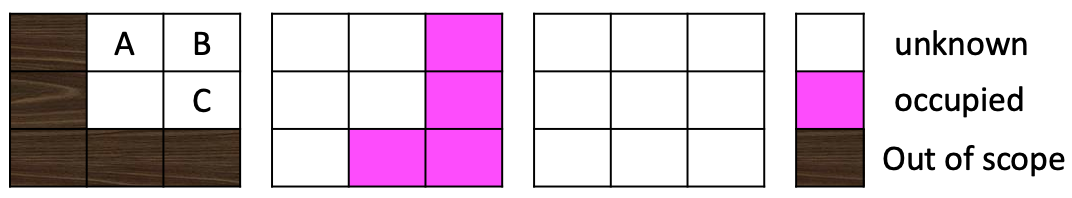}
    \caption{Special conditions when calculating cell boundariness}
    \label{fig:no_bd}
\end{figure}

However, several special cases need to be handled differently, as shown in Fig. \ref{fig:no_bd}.
\begin{itemize}
    \item Left: when a cell is close to the edge of the occupancy map, it has less than eight neighbors. 
    \item Middle: all known neighbors of a cell are occupied, then the cell is probably occluded. Its boundariness should be set to zero.
    \item Right: if all neighbors of a cell are unknown, the cell's boundariness should be set to zero. 
\end{itemize}

\begin{figure}
    \centering
    \vspace{2mm}
    \includegraphics[width =0.89\linewidth]{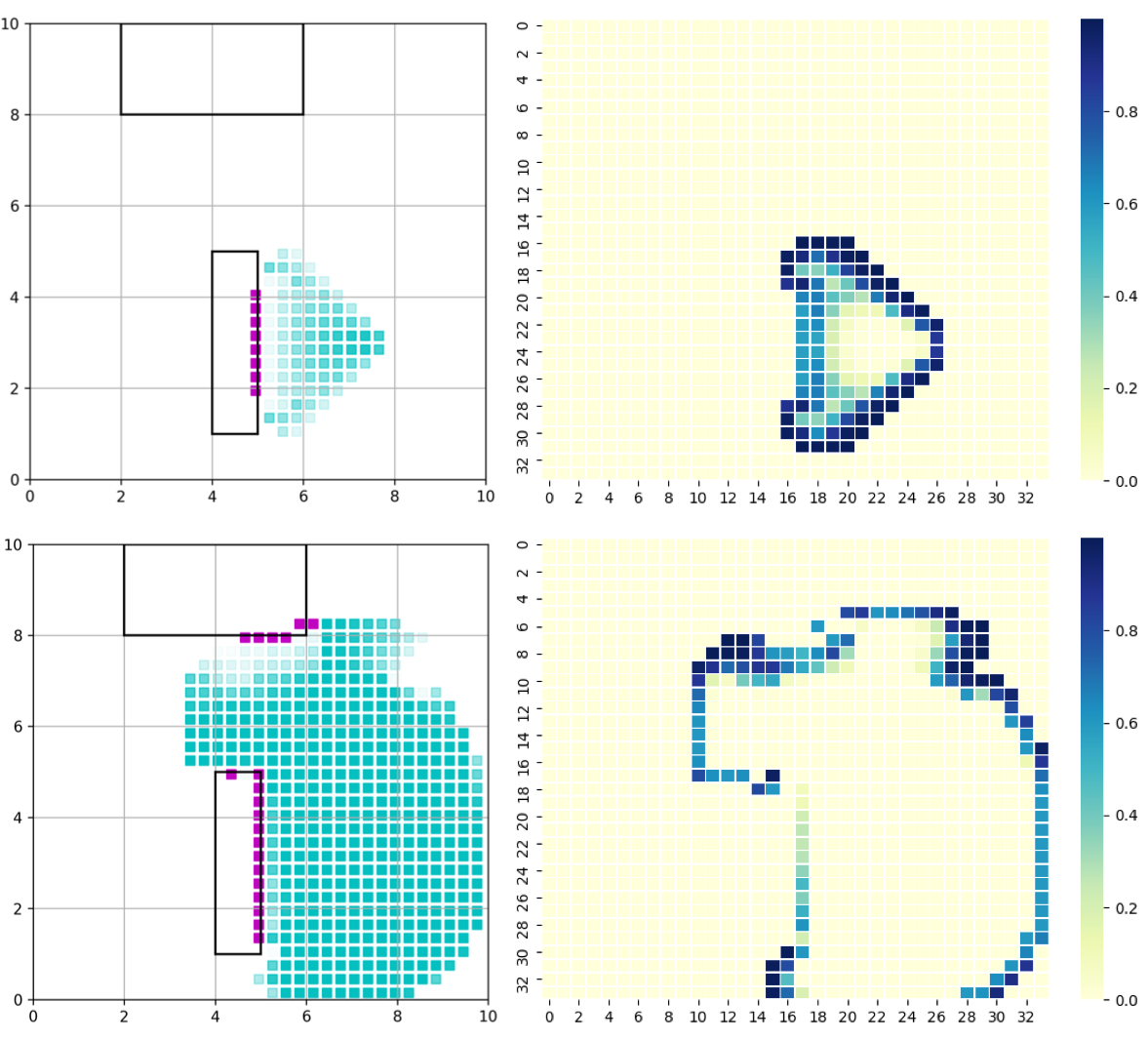}
    \caption{Occupancy maps (left) and their corresponding boundariness maps (right). The darker the color, the higher the boundariness.}
    \label{fig:boudnariness_map}
\end{figure}

Computing the boundariness of every cell in a probabilistic occupancy map, $\Map{odds}$, generates a boundariness map, $\Map{bd} \in \mathbb{R}^{m \times n}$. Examples of occupancy maps and their corresponding boundariness maps are shown in Fig. \ref{fig:boudnariness_map}.

{}

\vspace{-0.3cm}
\section{Information Gain}
\noindent Information gain is a measure of the number of boundary cells a view-point, $\xi$, or a sequence of view-points, $\Xi$, can observe. In many sampling-based exploration algorithms \cite{song2017online, bircher2016receding}, information gain is calculated by counting the number of boundary cells in the field of view of $\xi$ or $\Xi$. However, this computation is fundamentally discrete and ill-suited for optimization. In this section, we propose differentiable information gain functions for single and multiple view-points, which will be used for gradient-based optimizations in later sections.

\subsection{Fuzzy-Logic Filter}
\noindent The fuzzy-logic filter for a view-point $\xi$, $\Phi_{\xi}: \Lambda(\xi)\rightarrow [0, 1]$, is a
mapping from the set of cells surrounding $\xi$ to a value between 0 and 1 that measures how far away each cell is from the field of view of view-point $\xi$. The fuzzy-logic filter for a path $\Xi = [\xi_0, \xi_1, \dots \xi_k]$ can be defined analogously as $\Phi_{\Xi}: \Lambda(\Xi)\rightarrow [0, 1]$, where $\Lambda(\Xi)$ is a set of cells surrounding the path $\Xi$.

For a single view-point, $\xi = (x, y, \theta)$, $\Lambda(\xi)$ includes cells in a $4h \times 4h$ box with $\xi$ at the center of the box, where $h = \frac{R_{\text{max}}}{\varrho}$, and $\varrho$ is the cell resolution. For Cell $(i, j)$ in $\Lambda(\xi)$, its distance $\delta$ and direction $v$ relative to $\xi$ is calculated as:  
\begin{subequations}
\begin{align}
    \Vec{V} &= (i - 2h, j - 2h),\\
    \Vec{v}& =\Vec{V}/\|\Vec{V}\|_2, \\
    \delta &= \varrho\|\Vec{V}\|_2.
\end{align}
\end{subequations}
The direction of the robot, $\Vec{u}$, is given by $(\cos{\theta},\sin{\theta})$.

\begin{figure}[htp!]
 \vspace{-0.5cm}
    \centering
    \includegraphics[width = 0.98\linewidth]{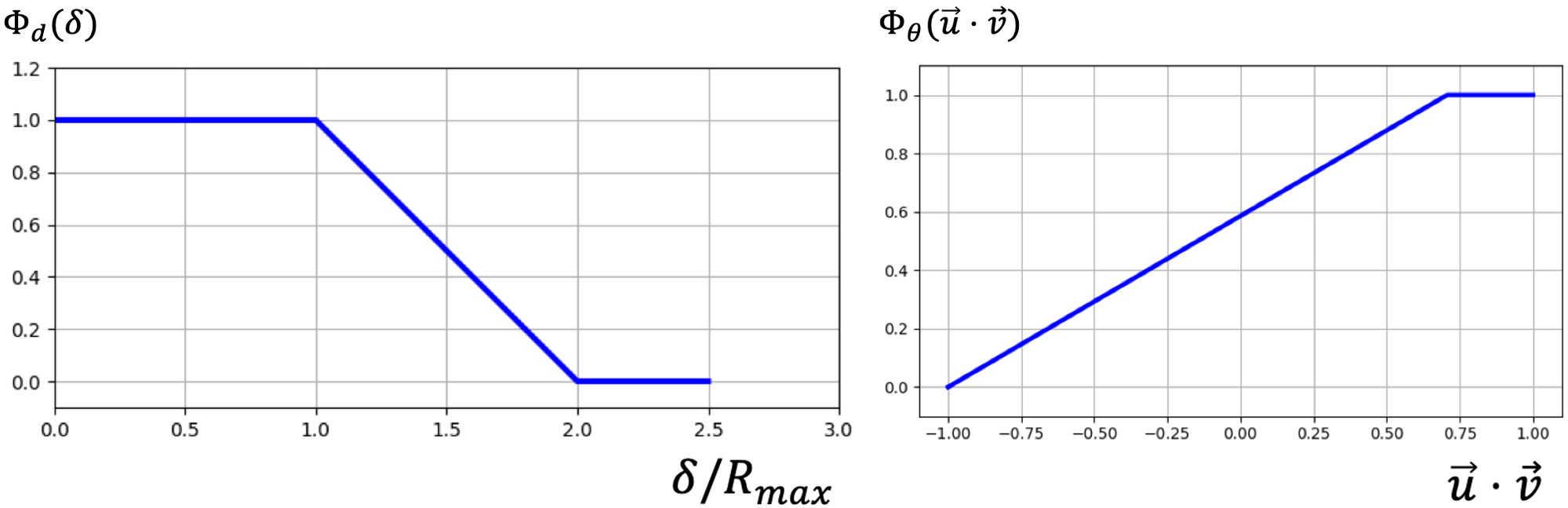}
    \caption{Discount factors for the distance and the angle}
    \label{fig.distance_angle_discount}
\end{figure}

As shown in Fig. \ref{fig.distance_angle_discount}, the discount values are the highest when a cell is inside the field of view and drop linearly to zero as the cell gets further away. The expressions for the discount values for distance $\phi_d$ and angle $\phi_{\theta}$ are: 
\begin{equation}
    \phi_d = \begin{cases}
  0 &  \delta > 2R_{\text{max}}\\    
  \frac{2R_{\text{max}} - \delta}{R_{\text{max}}} &  R_{\text{max}}\le \delta \le 2R_{\text{max}} \\
  1  &  \delta < R_{\text{max}}
\end{cases},
\end{equation}

\begin{equation}
\vspace{2mm}
\phi_\theta = \begin{cases}
\frac{1+\Vec{u}\Vec{v}}{1+\cos(\frac{\omega}{2})} & \Vec{u}\Vec{v} < \cos{\frac{\omega}{2}} \\
1 &  \Vec{u}\Vec{v} \geq \cos{\frac{\omega}{2}}
\end{cases}.
\end{equation}

Finally, the discount value of Cell $(i, j)$, $\Phi_{\xi}(i,j)$, is given as the product of the distance and angle discount values:
\begin{equation}
\Phi_{\xi}(i,j) = \phi_d(\delta)\phi_{\theta}(\Vec{u}, \Vec{v}, \omega).
\end{equation}

\begin{figure}[htp!]
\vspace{-0.3cm}
    \centering
    \includegraphics[width = 0.45\linewidth]{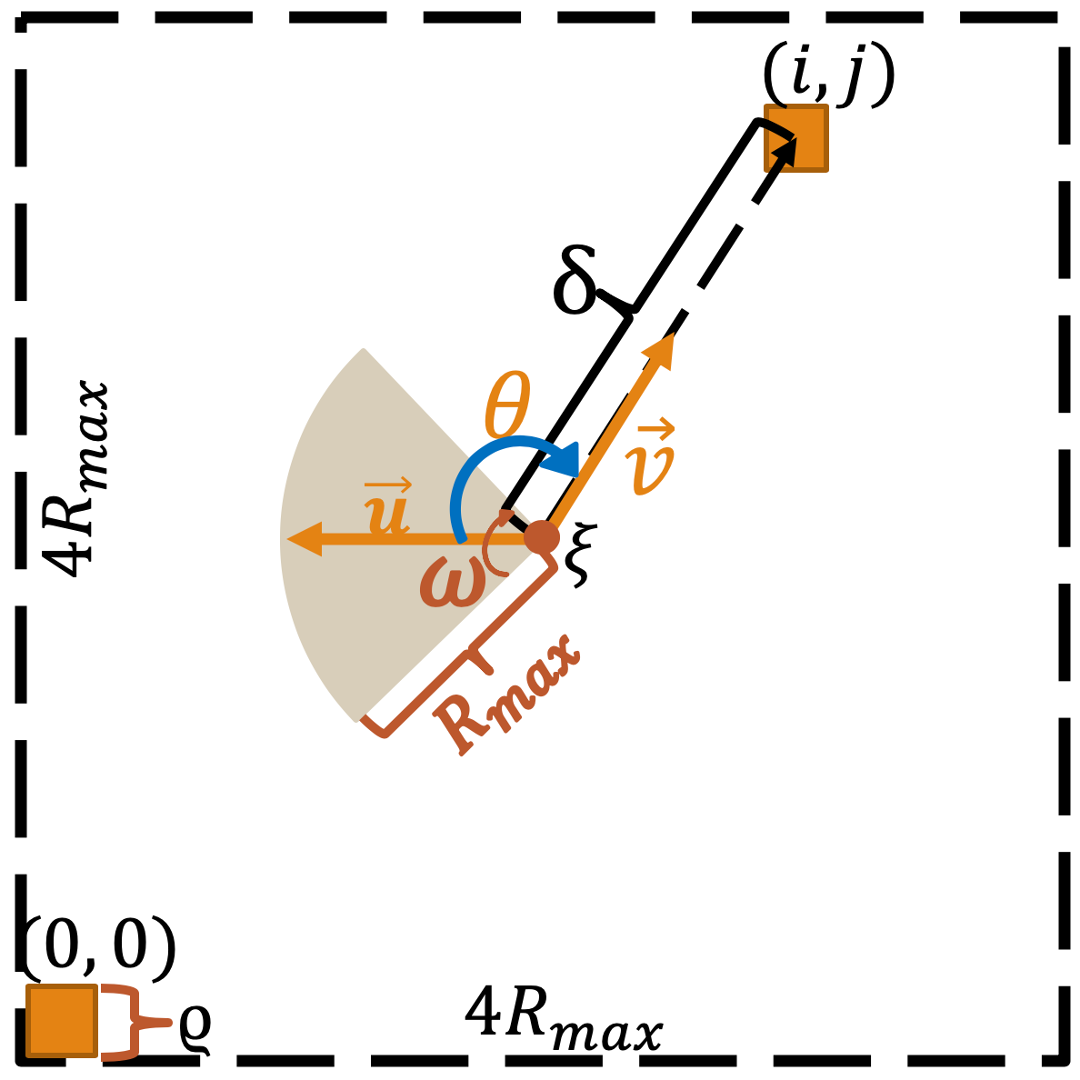}
    \caption{Calculate discount factor; the size of the region is a square with edge distance $4R_{\text{max}}$; shaded yellow area represents the robot's field of view; the distance and angle between sensor pose $\xi$ and Cell $(i,j)$ are $\delta$ and $\theta$; $\Vec{u}$ is the direction vector of the sensor, and $\Vec{v}$ is the vector from the sensor to Cell $(i,j)$.}
    \label{fig.discount}
\end{figure}

\begin{figure}[htp!]
    \vspace{0.1cm}
    \centering
    \includegraphics[width = 0.55\linewidth]{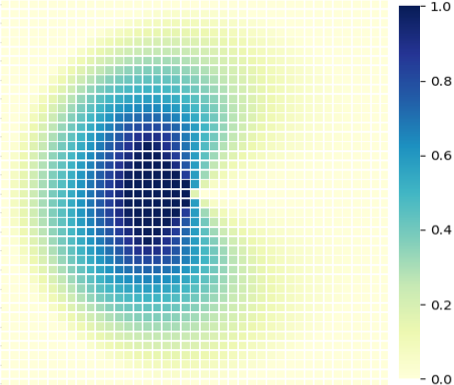}
    \caption{Fuzzy-logic filter for camera position in Fig. \ref{fig.discount}}
    \label{fig.discount_map}
    \vspace{-0.4cm}
\end{figure}

For the sensor position shown in Fig. \ref{fig.discount}, the corresponding fuzzy-logic filter looks like the one shown in Fig. \ref{fig.discount_map}.

{}

\subsection{Information Gain for View-Point $\xi$ \label{subsec.infogain_view}}
\noindent The information gain, IG, of a view-point, $\xi$, is a function of both $\xi$ and the current boundariness map $\Map{bd}$, i.e., $\mathrm{IG}_{text{view}}(\xi): \mathbb{R}^3 \times \mathbb{R}^{m \times n} \rightarrow \mathbb{R}$. It sums over all cells in $\Lambda(\xi)$ the product of each cell's discount value and boundariness:
\begin{equation}
\label{eqn.info_gain}
\mathrm{IG}_{\text{view}}(\xi, \mathbb{M}_{\text{bd}}) = \sum_{(i_{\mathbb{M}},j_{\mathbb{M}})\in \Lambda(\xi)}\Phi(g(i_{\mathbb{M}}, j_{\mathbb{M}})) \mathbb{M}_{\text{bd}}(i_{\mathbb{M}}, j_{\mathbb{M}}),
\end{equation}
where 
\begin{equation}
\label{eqn.map_index}
(i, j) = g(i_{\mathbb{M}}, j_{\mathbb{M}}) = (i_{\mathbb{M}} + 2h -  \lfloor\frac{\xi[0]}{\varrho}\rfloor, 
 j_{\mathbb{M}} + 2h - \lfloor\frac{\xi[1]}{\varrho}\rfloor)
\end{equation}
is the mapping from the boundary map index ($i_{\mathbb{M}}, j_{\mathbb{M}}$)  to the fuzzy-logic filter index ($i, j$).

\subsection{Information Gain for Path $\Xi$}
\begin{algorithm}[htp!]
\caption{Calculate Information Gain for a Path}
\label{alg.infogain_path}
\begin{algorithmic}[1]
\Require $\Xi$, $\Map{bd}$
\Ensure $f(\Xi)$
\State $L = \sum_{i=0}^{k-1}\|\xi_{i+1}-\xi_i\|_2$, $N = \lceil \frac{L}{2\varrho}\rceil$
\If{k $<$ N}
\State $\Xi_s$ $\gets$ Sample($\Xi$, $N - k$)
\EndIf
\State $\Xi = \left(\Xi\bigcup\Xi_s\right)\setminus\{\xi_0, \xi_k\}$
\State $\Lambda(\Xi)=\bigcup_{\xi_i\in\Xi} \Lambda(\xi_i) \setminus{\bigcup_{\xi_j \in \{\xi_0, \xi_k\}}}\Lambda_{\xi_j}$
\State $f(\Xi) = \sum_{(i_{\mathbb{M}},j_{\mathbb{M}})\in \Lambda(\Xi)}\Phi_\Xi(i_{\mathbb{M}},j_{\mathbb{M}}) \mathbb{M}_{\text{bd}}(i_{\mathbb{M}}, j_{\mathbb{M}})$
\end{algorithmic}
\end{algorithm}

Path $\Xi$ is a collection of view-points: $\Xi = \{\xi_0, \xi_1, \dots, \xi_k\}$. Since the view regions of vertices in $\Xi$ may overlap, we cannot compute the information gain of the path by directly summing information gains of individual vertices. Thus, we create a dynamic dictionary of fuzzy-logic filter $\Phi_\Xi$ for the path with $(i_{\mathbb{M}},j_{\mathbb{M}})$ in Eqn. (\ref{eqn.map_index}) as keys. This way, we can iteratively add or update the values in $\Phi_\Xi$ when calculating the fuzzy-logic filter for each view-point. 

Alg. \ref{alg.infogain_path} shows the procedure of calculating the path information gain using $\Phi_\Xi$. $N$ in Line 2 is the expected number of view-points along a path with length $L$. If the total number of vertices in the path is less than $N$, we randomly sample $N - k$ points on the path in Line 5. As both the start and end points in the path are fixed, the goal is to maximize the boundary cells only covered by the intermediate view-points as shown in Lines 6 and 7. Similar to Eqn. (\ref{eqn.info_gain}), information gain of the path is calculated in Line 8.

\section{Gradient-based Path Optimization}

\noindent This section introduces a gradient-based path optimization method which tries to improve Path $\Xi$'s smoothness and information gain simultaneously. A sampling-based method for creating an initial guess for the optimization is also introduced. 

\subsection{Objective Function}
\noindent Given a path, $\Xi_0 = [\xi_0, \xi_1, \dots \xi_k]$, we would like to improve both its smoothness and information gain by minimizing the following objective function:
\begin{equation}
\label{eqn:objective_function}
    f(\Xi) = \alpha \sum_{i = 1}^k \norm{\xi_i - \xi_{i-1}}_{\bm{W}}^2 -  \mathrm{IG}_{\text{path}}(\Xi, \Map{bd}).
\end{equation}
The first term penalizes non-smoothness in the path. $\alpha$ is the weight of this term. $\norm{\cdot}_{\bm{W}}$ represents the norm induced by symmetric positive definite matrix $\bm{W}$, i.e., $\norm{\bm{v}}^2_{\bm{W}} = \bm{v}^{\intercal} \bm{W} \bm{v}$. We choose $\bm{W} = \mathrm{diag}([1, 1, 0.1])$ so that change in the orientation, $\theta$, is less costly compared to change in position, $x$ and $y$. There is a minus sign before the path information gain, $\mathrm{IG}_{\text{path}}$, because it is a quantity which we would like to maximize.

\subsection{Initial Guess}
\noindent The initial guess, $\Xi_0 = [\xi_0, \xi_1, \dots \xi_k]$, is generated using RRT and then shortened with a common smoothing technique \cite{lavalle2006planning}. Our observation is that seeding the optimizer using a path generated by the sampling-based method helps the optimizer avoid local minima and achieve more complete coverage. The starting view-point,  $\xi_0$, is simply the current configuration of the robot. The goal view-point, $\xi_k$, is vital for guiding the exploration so that we will elaborate on the goal-selection method.

There are three requirements for selecting the goal, $\xi_k$: (1) $\xi_k$ covers as many boundary cells as possible; (2) $\xi_k$ does not get too close to obstacles; (3) the distance between $\xi_0$ and $\xi_0$ is small. Accordingly, the scoring function, $f_\text{score}: \mathbb{R}^3 \times \mathbb{R}^{m \times n} \rightarrow \mathbb{R}$, which takes a view-point, $\xi$, and the current boundariness map, $\Map{bd}$, as its input and returns a scalar score, is defined as:
\begin{equation}
\label{eqn.score_goal}
    f_{\text{score}}(\xi, \mathbb{M}_{\text{bd}}) = \sum_{(i,j)\in \Lambda(\xi)}\mathbb{M}_{\text{bd}}(i,j)e^{-\lambda _1\delta}e^{-\lambda_2N_{\text{occupied}}},
\end{equation}
where $\Lambda(\xi)$ is the set of cells in the field of view of the robot at view-point $\xi$; $\lambda_1, \lambda_2 \in (0,1)$ are constant weights; $\delta = \norm{\xi_0 - \xi_k}$; $N_{\text{occupied}}$ is the number of obstacles in a user-defined box centered at $\xi$. The goal, $\xi_k$, is selected by sampling view-points on the entire grid, scoring them using the function in Eqn. (\ref{eqn.score_goal}), and picking the one with the highest score as $\xi_k$.

\subsection{Results}
\begin{figure}[htp!]
    \centering
    \subfloat[NBV-RRT]{
    \includegraphics[width=0.45\linewidth]{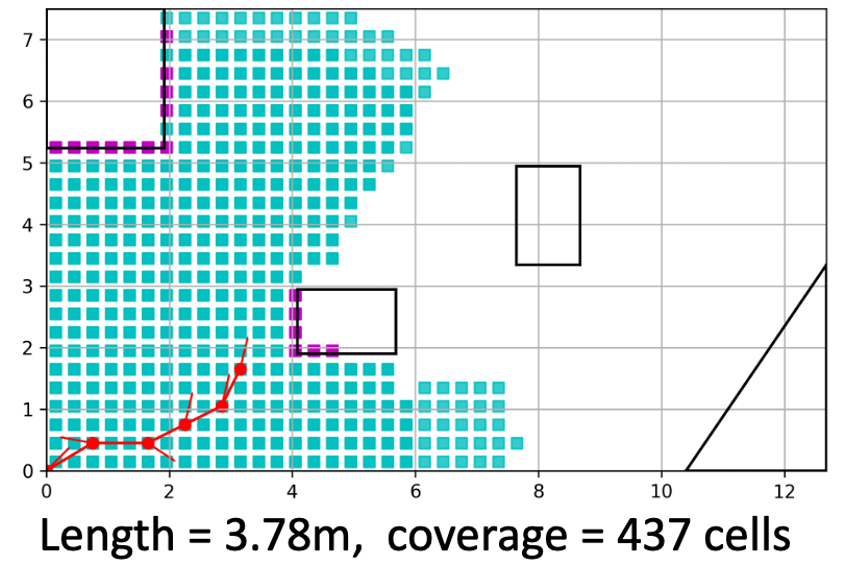}
    }   
    \subfloat[NBV-RRT with Optimization]{
    \includegraphics[width=0.45\linewidth]{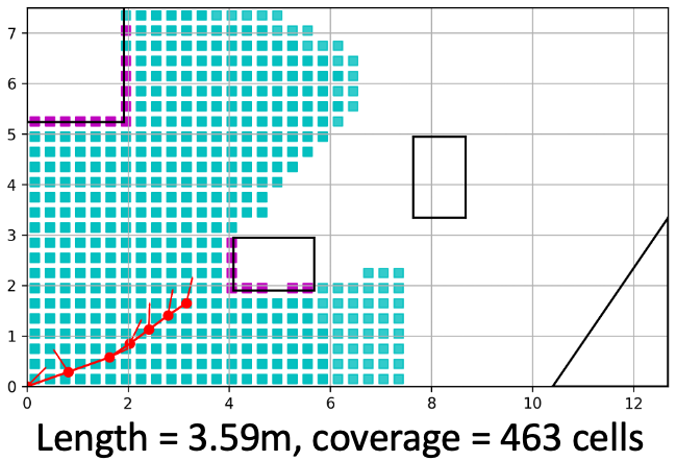}
    }   
    \caption{Path optimization through a gradient descent method. Red dots with arrows represent the positions and orientations of the robot.}
    \label{fig.gd}
    \vspace{-0.3cm}
\end{figure}

\noindent The gradient of the differentiable objective function defined in Eqn. (\ref{eqn:objective_function}) can be evaluated using back-propagation \cite{griewank1989automatic}. Fig. \ref{fig.gd} is an example of applying gradient descent to a path generated by a sampling-based planner. The gradient-based optimization reduces the path length from 3m to 1.95m and increases coverage by 15\%. 

\section{Gradient-Based Exploration}

\noindent Alg. \ref{algo:gradient_planning} is an overview of the gradient-based path planning algorithm to determine the occupancy map of the environment. The inputs to the system are: (1) the robot's initial configuration, $\xi_0$, (2) the onboard sensor model with a maximum range of $R_{\text{max}}$ and a beam aperture of $\omega$, (3) the dimension of the grid map $(m, n)$ with cell resolution $\varrho$, and (4) the number of points to be sampled for goal selection $N_{s}$.  

The log-odds map, $\mathbb{M}_{\text{odds}}$, and the boundariness map, $\mathbb{M}_{\text{bd}}$, are initialized as $m \times n$ matrices with all elements equal to zero. $f_\Xi$ is defined as the information gains for $\tau_2$ consecutive paths, and it is initialized as a list of $\textbf{1}$s with $\tau_2$ elements. Line 2 updates both $\mathbb{M}_{\text{odds}}$ and $\mathbb{M}_{\text{bd}}$ based on the measurement from the initial configuration, $\xi_0$. If the number of cells in $\mathbb{M}_{\text{bd}}$ with probabilities of being at the boundary is greater than the constant threshold, $\tau_1$, and the summation of the information gains over the last $\tau_2$ consecutive paths is greater than zero, we start to explore the environment. 

Exploration path generation within the loop contains three main steps: (1) initial path generation with a greedy method with Lines 4 and 5; (2) path optimization through gradient descent in Line 6; (3) map revolution with a designed path from Line 8 to 11. Then in Line 12, we deploy the robot to the endpoint in the path and repeat the previous steps if all requirements in Line 3 are satisfied. After exploration, this algorithm calculates probability map $\mathbb{M}_{\text{prob}}$ from log-odds map $\mathbb{M}_{\text{odds}}$ to understand the distribution of obstacles and free space in the environment.

\begin{algorithm}[htp!]
\caption{Gradient-Based Area Coverage}
\label{algo:gradient_planning}
\begin{algorithmic}[1]
\Require $\xi_{0}$, $R_{\text{max}}$, $\omega$, m, n, $\varrho$, $N_{s}$
\Ensure $\mathbb{M}_{\text{prob}}$
\State $\mathbb{M}_{\text{odds}}=\mathbb{M}_{\text{bd}} = 0_{m,n}$,$f_\Xi = 1_{\tau_2}$
\State $\mathbb{M}_{\text{odds}}, \mathbb{M}_{\text{bd}} \gets$ UpdateMap($\xi_0$, $R_{\text{max}}$, $\omega$, $\varrho$, $\mathbb{M}_{\text{odds}}$, $\mathbb{M}_{\text{bd}}$)
\While{size($\mathbb{M}_{\text{bd}} > \rho$)$>\tau_1$ $\textbf{and}$ $\sum(f_\Xi) \neq 0$}
\State $\xi_{\text{goal}} \gets$ FindGoal($\mathbb{M}_{\text{odds}}$, $\mathbb{M}_{\text{bd}}$, $N_{s}$, $R_{\text{max}}$, $\omega$, $\varrho$)
\State $\Xi \gets$ RRTPlanner($\xi_{0}$, $\xi_{\text{goal}}$, $R_{\text{max}}$, $\omega$, $\mathbb{M}_{\text{odds}}, \varrho)$
\State $\Xi \gets$ GradientPlanner($\Xi, \mathbb{M}_{\text{odds}}, \mathbb{M}_{\text{bd}}$)
\State $f_\Xi \gets$ UpdateInfoGain($\Xi, \mathbb{M}_{\text{bd}}, \varrho, R_{\text{max}}$) 
\For{$\xi \in \Xi$}
\State $\mathbb{M}_{\text{odds}}, \mathbb{M}_{\text{bd}} \gets$ \\ \hspace{2cm} UpdateMap($\xi$, $R_{\text{max}}$, $\omega$, $\varrho$, $\mathbb{M}_{\text{odds}}$, $\mathbb{M}_{\text{bd}}$)

\EndFor
\State $\xi_0=\Xi[-1]$
\EndWhile
\State $\mathbb{M}_{\text{prob}} \gets$ GetProbMap($\mathbb{M}_{\text{odds}}$)
\end{algorithmic}
\end{algorithm}

\section{Results and Discussion}
\begin{figure}[htp!]
    \vspace{-0.3cm}
    \centering
        \subfloat[Information gain]{
    \includegraphics[width=0.46\linewidth]{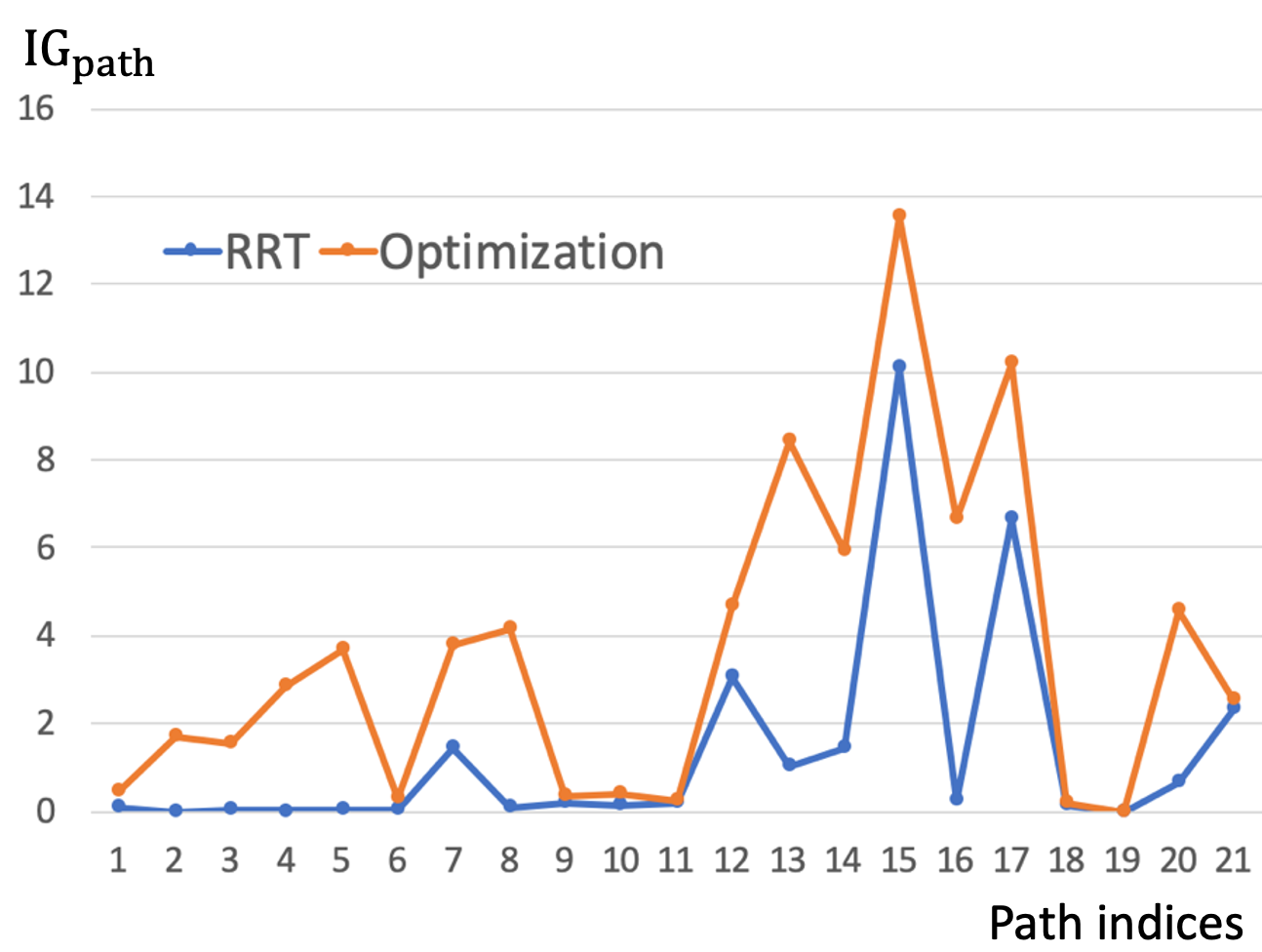}
    }   
    \subfloat[Path length]{
    \includegraphics[width=0.46\linewidth]{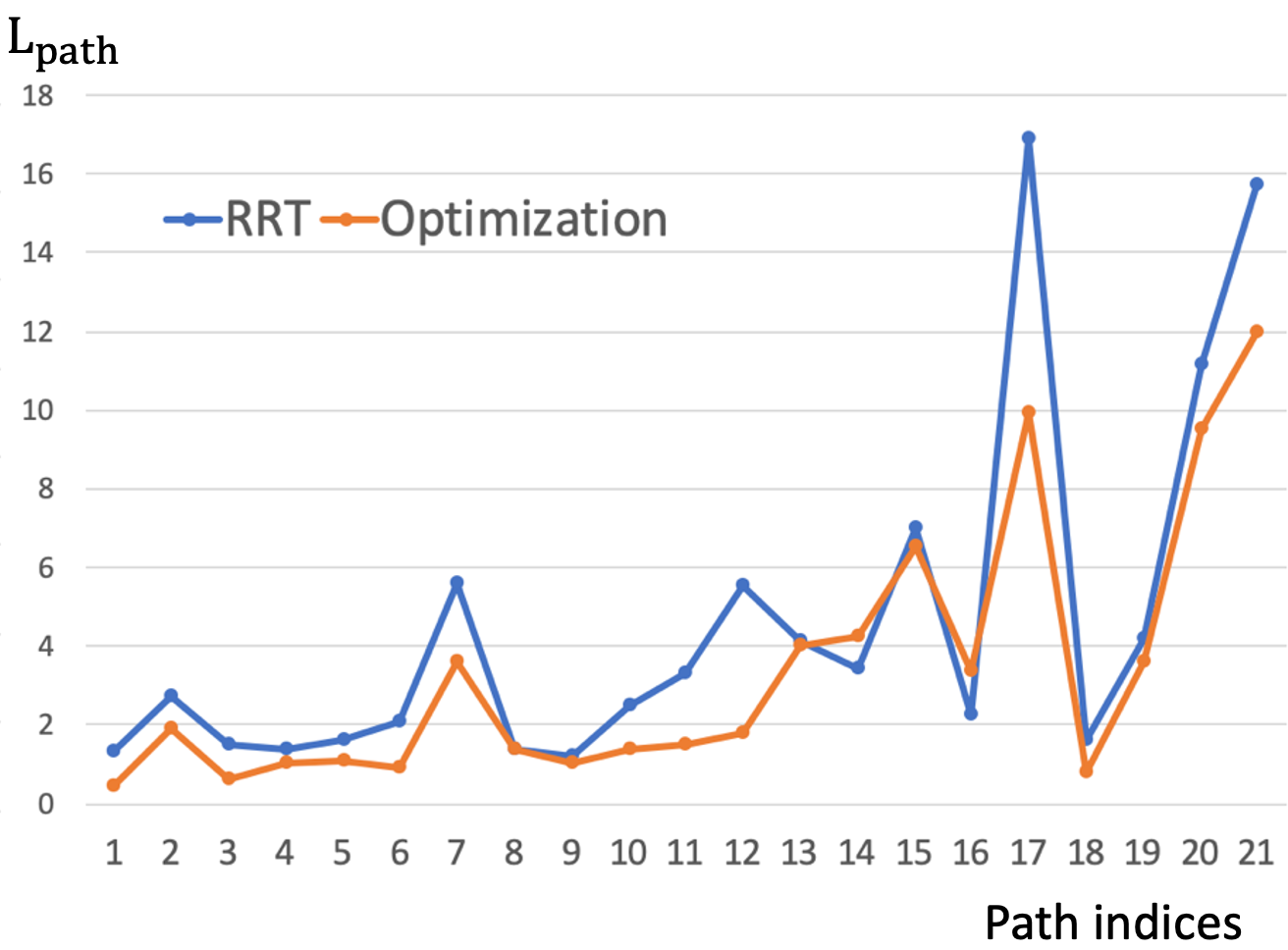}
    }   
    \caption{Comparison between RRT and optimization}
    \label{fig.rrt_optimization}
\end{figure}

\noindent Simulations and experiments with a Turtlebot3 Burger robot prove that our optimization algorithm improves area coverage and significantly reduces path length. As shown in Fig. \ref{fig.rrt_optimization}, we compare the area coverage and path length before and after executing the optimization algorithm. After iteratively executing the proposed path optimization algorithm 21 times, the robot covers 98.5\% of a region, consisting of 32 × 32 cells with a resolution of 0.3m. It has an average of 173\% improvement of area coverage and 37\% reduction of path length.

The log-odds upper and lower bounds used in the simulations are:
\begin{equation}
    l_{\text{max}} = \log\frac{0.9}{0.1}, l_{\text{min}} = \log\frac{0.3}{0.7}. 
\end{equation}

{}

\begin{figure}[htp!]
    \vspace{0.2cm}
    \centering
    \subfloat[NBV-RRT with smoothing]{
\includegraphics[width=0.4\linewidth]{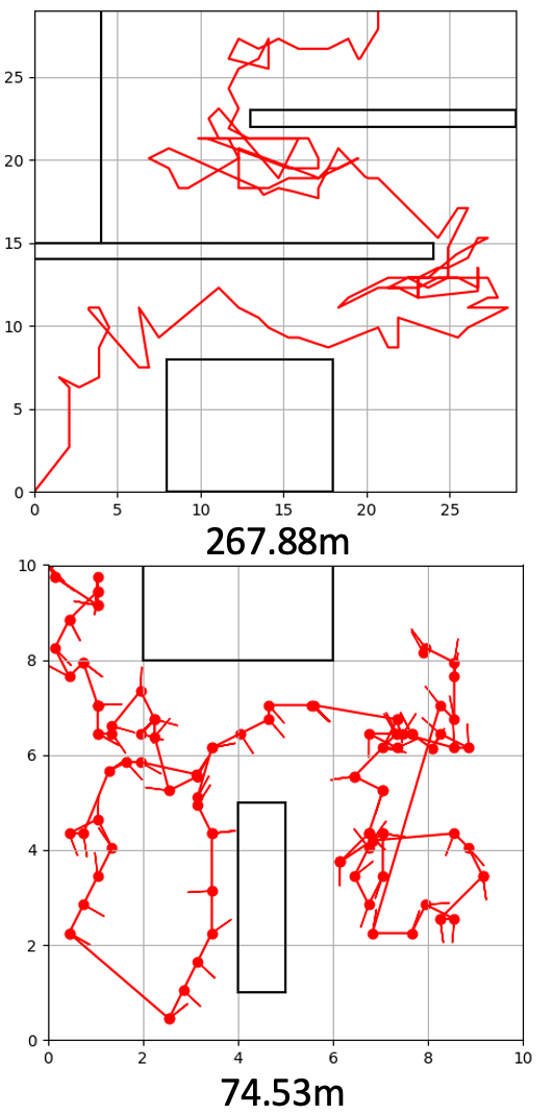}
}   
\subfloat[NBV-RRT with optimization]{
\includegraphics[width=0.4\linewidth]{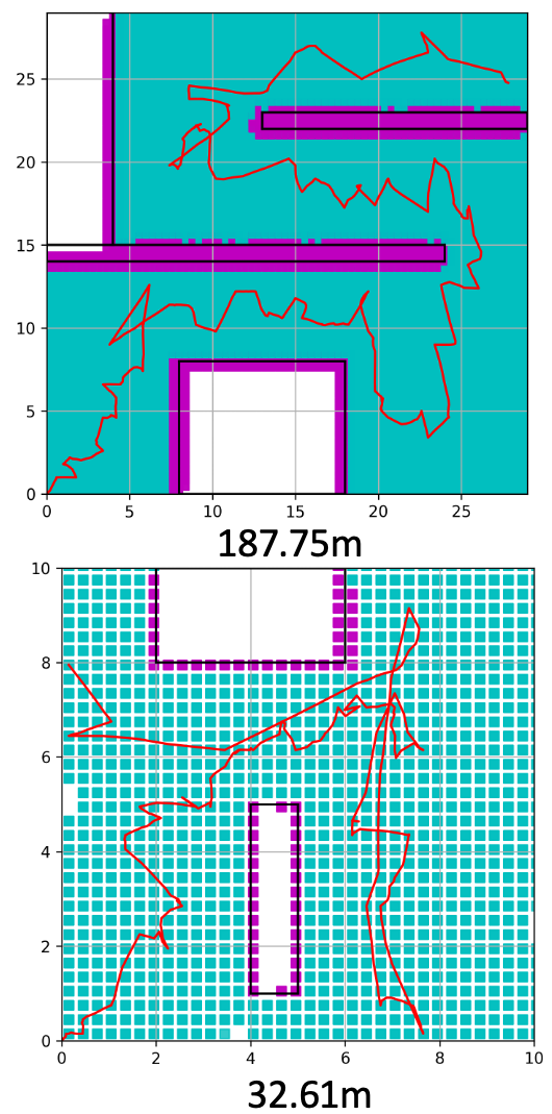}
}   
    \caption{Comparison of path lengths between RRT-smoothing algorithm and RRT-gradient descent algorithm}
    \label{fig:my_result}
    \vspace{-0.1cm}
\end{figure}

Our planner has consistent performance in different maps, as shown in Fig. \ref{fig:my_result}. In these two cases, the lengths of paths are reduced by 43.7\% and 40.9\%  without sacrificing the coverage area.

{}

\begin{figure}[h]
    \centering
    \vspace{4mm}
    \includegraphics[width = 0.98\linewidth]{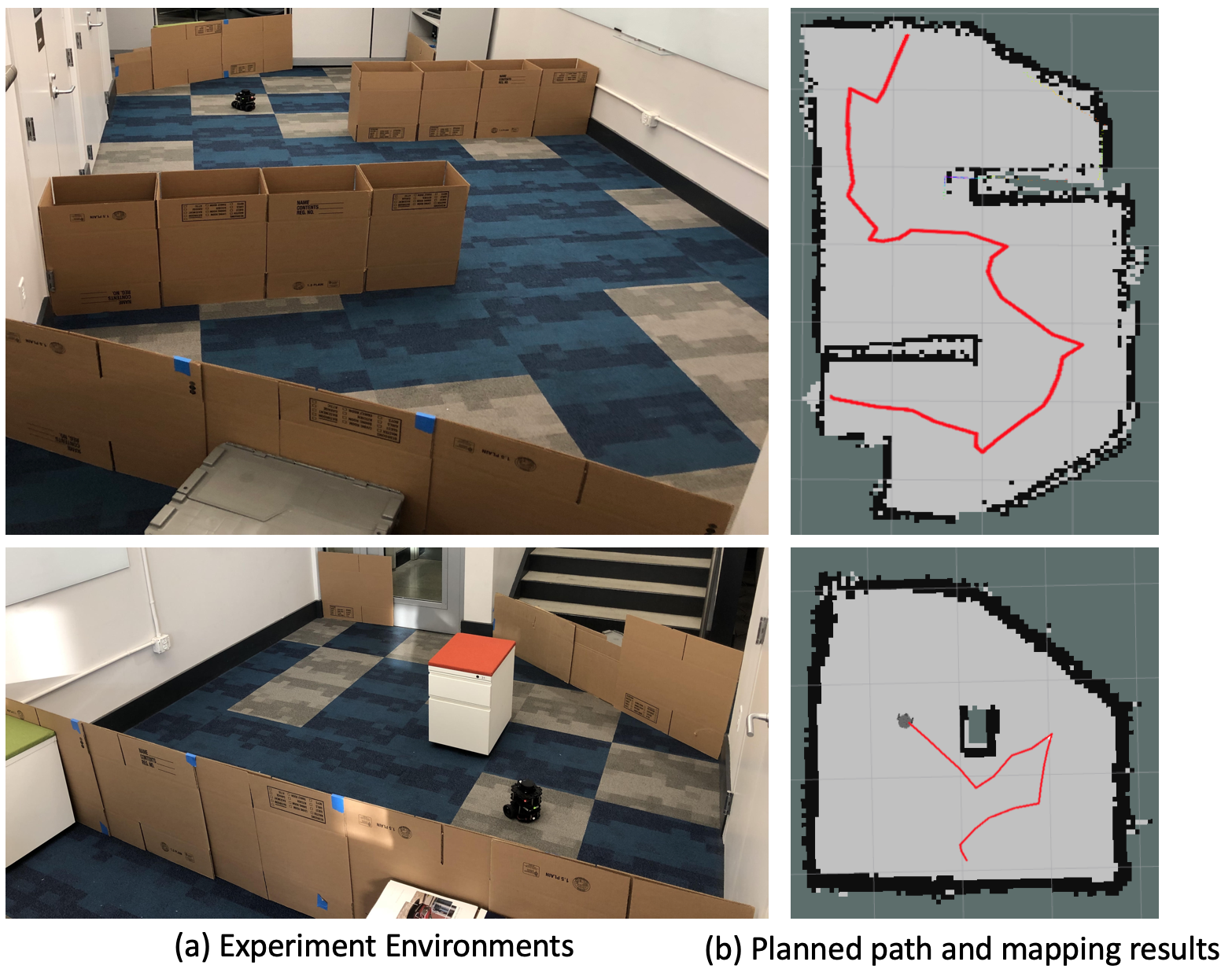}
    \caption{Experiment results}
    \label{fig:exp1}
\end{figure}

We then test our algorithm on a Turtlebot3 Burger robot in closed regions with a size of 6.5m $\times$ 3.5m and 3.0m $\times$ 3.5m. The onboard 2D laser scanner has FOV ($\omega$) of 90 degrees, maximum view range ($R_{\text{max}}$) of 3m, and sensor resolution of 1$\degree$.  The results shown in Fig. \ref{fig:exp1} demonstrate that our optimization-based planner generates relatively smooth paths and successfully covers the target regions.

{}

\section{Conclusions}
\noindent We proposed a novel algorithm for the optimizing exploration paths of a robot to cover unknown 2D areas. This algorithm starts with a path planned with the Next-Best-View algorithm, evaluates both the information gain of uniformly sampled view-points along the path, and then formulates a multi-objective utility function to maximize the coverage of boundary cells of these view-points through a gradient-descent method. To quantitatively measure the quality of the space being explored and assess a view-point's inspectability, we introduced a boundariness map, $\mathbb{M}_{\text{bd}}$, that estimates the likelihood of a cell being a boundary cell. Furthermore, the planner generates a path iteratively based on a pre-defined frequency to avoid unnecessary movements. Simulations and experimental results show that our algorithm improves area coverage while reducing path lengths.      

\addtolength{\textheight}{-12cm}   

\section*{Acknowledgement}
\noindent The authors would like to thank TOPRISE Co., LTD. for their financial support for this work.
\bibliographystyle{IEEEtran}
\bibliography{main}

\end{document}